%% file: main.tex
\documentclass{article}

\usepackage[main, final]{neurips_2026}

\usepackage[utf8]{inputenc} 
\usepackage[T1]{fontenc}    
\usepackage{hyperref}       
\usepackage{url}            
\usepackage{booktabs}       
\usepackage{amsfonts}       
\usepackage{nicefrac}       
\usepackage{microtype}      
\usepackage{xcolor}         

\title{Large Discrete Policy: Advancing Explicit Behavior Modeling with Stochastic Iterative Scoring}

\author{
\textbf{Zhenxin Li$^{1,2*}$, Nadine Chang$^{3}$, Xinglong Sun$^{3}$, Jingde Chen$^{3}$, Wenhao Yao$^{1,2}$, Zi Wang$^{3}$} \\
\textbf{Maying Shen$^{3}$, Yu-Gang Jiang$^{1,2\dagger}$, Zuxuan Wu$^{1,2\dagger}$, Shiyi Lan$^{3}$, Jose M. Alvarez$^{3}$} \\
\\
$^1$Institute of Trustworthy Embodied AI, Fudan University \\
$^2$Shanghai Key Laboratory of Multimodal Embodied AI \quad $^3$NVIDIA\\
}

\usepackage{xcolor}
\definecolor{citeblue}{RGB}{0, 114, 189} 

\usepackage{hyperref}
\hypersetup{
    colorlinks=true,    
    citecolor=citeblue, 
    linkcolor=red,      
    urlcolor=blue       
}

\usepackage{graphicx}
\usepackage{booktabs}
\usepackage{xspace}
\usepackage[dvipsnames]{xcolor}
\usepackage{color, colortbl}

\usepackage{amsmath}
\usepackage{makecell}
\usepackage{multirow}
\usepackage{float}
\usepackage{amssymb}
\usepackage{utfsym}
\usepackage{diagbox}
\usepackage{dsfont}
\usepackage{enumitem}
\PassOptionsToPackage{numbers, compress}{natbib}
\newcommand{\hydra}{LDiP\xspace}

\begin{document}

\maketitle

\begin{abstract}
\input{sections/0_abs}
\end{abstract}

\input{sections/1_intro}

\input{sections/2_related}

\input{sections/3_method}
\input{sections/4_exp}
\input{sections/5_conclusion}
\input{sections/7_ack}

\bibliographystyle{plain} 
\bibliography{main}


\newpage
\appendix
\input{sections/6_appendix}





\end{document}

%% file: sections/0_abs.tex
Behavior policies are often formulated as continuous generative models, whose iterative denoising processes are expressive but difficult to interpret and prone to producing implausible actions. We propose the Large Discrete Policy (\hydra{}), a fully discrete behavior modeling framework that selects actions from a large vocabulary of physically plausible candidates. Rather than perturbing actions, \hydra{} improves expressivity through stochastic iterative scoring: it progressively re-scores and prunes candidates with score-space stochasticity, enabling fine-grained ranking and exploration among plausible actions while preserving an explicit decision process. Across end-to-end planning, closed-loop driving, robotic manipulation, and vision-language-action settings, \hydra{} consistently outperforms strong discrete and continuous baselines in autonomous driving, and exceeds or matches continuous generative policies in robotic manipulation. These results show that discrete policies, when equipped with effective scoring mechanisms, offer an expressive, plausible, and interpretable alternative for behavior modeling. Project website: \url{https://zhenxinli.net/LargeDiscretePolicy/}.

\newcommand\blfootnote[1]{%
  \begingroup
  \renewcommand\thefootnote{}\footnote{#1}%
  \addtocounter{footnote}{-1}%
  \endgroup
}
\blfootnote{$^{*}$Work done during an internship at NVIDIA. $^{\dagger}$Corresponding authors.}

%% file: sections/1_intro.tex
\section{Introduction}

Behavior learning is a central problem in physical AI systems, with applications ranging from autonomous driving~\cite{hu2023planning} to robotic manipulation~\cite{actzhao, chi2025diffusion}. 
The goal is to learn a policy that models behavior conditioned on visual observations, which is commonly achieved by leveraging offline demonstrations from expert agents~\cite{torabi2018behavioral}.

Since the action space in the physical world is typically continuous, behavior policies are commonly formulated as continuous models, including direct regression~\cite{hu2023planning, actzhao} and generative models~\cite{ho2020denoising, chi2025diffusion, lipman2023flow, black2024pi_0}. Recently, iterative denoising models (e.g. Diffusion Models~\cite{ho2020denoising, chi2025diffusion}, Flow-Matching Models~\cite{lipman2023flow, black2024pi_0}) demonstrate strong capacity for modeling multimodal behavior distributions~\cite{chi2025diffusion}, making them a popular choice in both autonomous driving~\cite{liao2025diffusiondrive} and robotics~\cite{chi2025diffusion, black2024pi_0}. Nevertheless, their generation process is largely implicit: actions emerge through repeated refinement of noisy samples, making it difficult to interpret how a behavior is formed, as illustrated in Fig.~\ref{fig:teaser} (a). Moreover, errors in the denoising process may lead to physically implausible actions~\cite{yang2025fine}.

Discrete policies provide an alternative. They are widely adopted in language modeling~\cite{vaswani2017attention}, where the action space is a finite vocabulary of symbols. Similar ideas have also been explored for behavior modeling, including trajectory prediction in autonomous driving~\cite{philion2020lift, shafiullah2022behavior, chen2024vadv2, li2024hydra, li2025generalized} and robotic manipulation~\cite{shafiullah2022behavior}.
Instead of generating continuous actions, these methods explicitly select from a discrete action vocabulary, as shown in Fig.~\ref{fig:teaser} (b). To reduce the discretization gap, prior work either predicts continuous offsets from selected candidates~\cite{shafiullah2022behavior}, or constructs dense action vocabularies~\cite{philion2020lift} with physically plausible actions. The latter approach, which we refer to as fully discrete policies, assigns scores to each action candidate.
By evaluating all candidates explicitly, it provides better interpretability than the implicit denoising process.

Despite these advantages, fully discrete policies still suffer from limited expressivity. A common concern is that a fixed action vocabulary may be too rigid to cover all behavior modes. To address this, existing discrete policies augment candidates with action-level perturbations~\cite{yao2026had}. While effective, such perturbations blur the fully discrete formulation and may produce physically implausible actions~\cite{yang2025fine, zheng2025diffusionbased, yao2026had}. On the other hand, recent work suggests that a sufficiently dense action vocabulary can already approximate continuous behavior spaces well~\cite{sun2026sparsedrivev2}. This motivates a different hypothesis: the bottleneck may not be the lack of feasible action candidates, but the difficulty of accurately scoring and ranking many near-optimal candidates. Prior methods improve ranking with additional refinement modules~\cite{yao2025drivesuprim}, but they leave the key question unanswered: can we improve the expressivity of discrete policies with an advanced, general scoring mechanism, while keeping the action vocabulary fixed and physically plausible throughout the process?

\input{figures/teaser_}

We propose the \textbf{Large Discrete Policy} (\hydra), which establishes fully discrete action scoring as a general and expressive framework for behavior modeling.
\hydra{} builds upon an advanced scoring mechanism, which preserves the explicitness and plausibility of discrete policies while substantially improving the behavior modeling ability.

To realize this goal, \hydra{} first constructs a large action vocabulary by clustering offline demonstrations~\cite{philion2020lift, li2024hydra}. To facilitate fine-grained comparisons among near-optimal candidates, \hydra{} performs \emph{iterative scoring}: action candidates are progressively pruned according to their scores until a single executable action remains, as shown in Fig.~\ref{fig:teaser} (c). This iterative process makes the decision process explicit: it reveals which candidates are considered, how they are ranked, and how the final action is selected.
To further address uncertainty and approximation errors in the scoring model~\cite{sima2025centaur}, we introduce \emph{stochastic scoring}, a novel mechanism that perturbs the scores during training rather than injecting noise into actions, which encourages exploration among plausible candidates while preserving the fully discrete formulation.

We benchmark \hydra{} on various behavior modeling tasks, including end-to-end planning for autonomous driving~\cite{dauner2024navsim, cao2025pseudo}, simulated closed-loop driving~\cite{zhou2024hugsim}, and robotic manipulation~\cite{Mandlekaretal2022}. 
\hydra{} consistently outperforms both discrete and continuous baselines in autonomous driving, while achieving performance stronger than or on par with the standard Diffusion Policy~\cite{chi2025diffusion} in high-DoF manipulation tasks.
Moreover, we extend \hydra{} by integrating it with Vision-Language Models (VLMs), yielding a Vision-Language-Action variant, \textbf{\hydra{}-VLA}. \hydra{}-VLA outperforms prior approaches, including diffusion-based~\cite{li2025recogdrive} and flow-based~\cite{zhou2026spanvla} VLAs.
Our main contributions are summarized as follows:

\begin{itemize}[
    label=\textbullet,
    topsep=0.2\baselineskip,
    itemsep=0.2\baselineskip,
    parsep=0pt,
    leftmargin=*
]
\item We introduce the \textbf{Large Discrete Policy} (\hydra{}), a fully discrete behavior modeling framework. \hydra{} employs an iterative scoring process over a large action vocabulary, together with a novel stochastic scoring mechanism. These designs preserve the fully discrete formulation while enhancing policy expressivity, enabling both physically plausible and interpretable behavior modeling.

\item We benchmark \hydra{} across end-to-end planning, closed-loop driving, and robotic manipulation, where it consistently outperforms strong discrete and continuous baselines, and matches or exceeds diffusion-based policies in high-DoF tasks. When extended with Vision-Language models, \hydra{}-VLA outperforms recent diffusion- and flow-based VLAs.
\end{itemize}

%% file: figures/teaser_.tex
\begin{figure*}[!t]
    \centering
    \includegraphics[width=\linewidth]{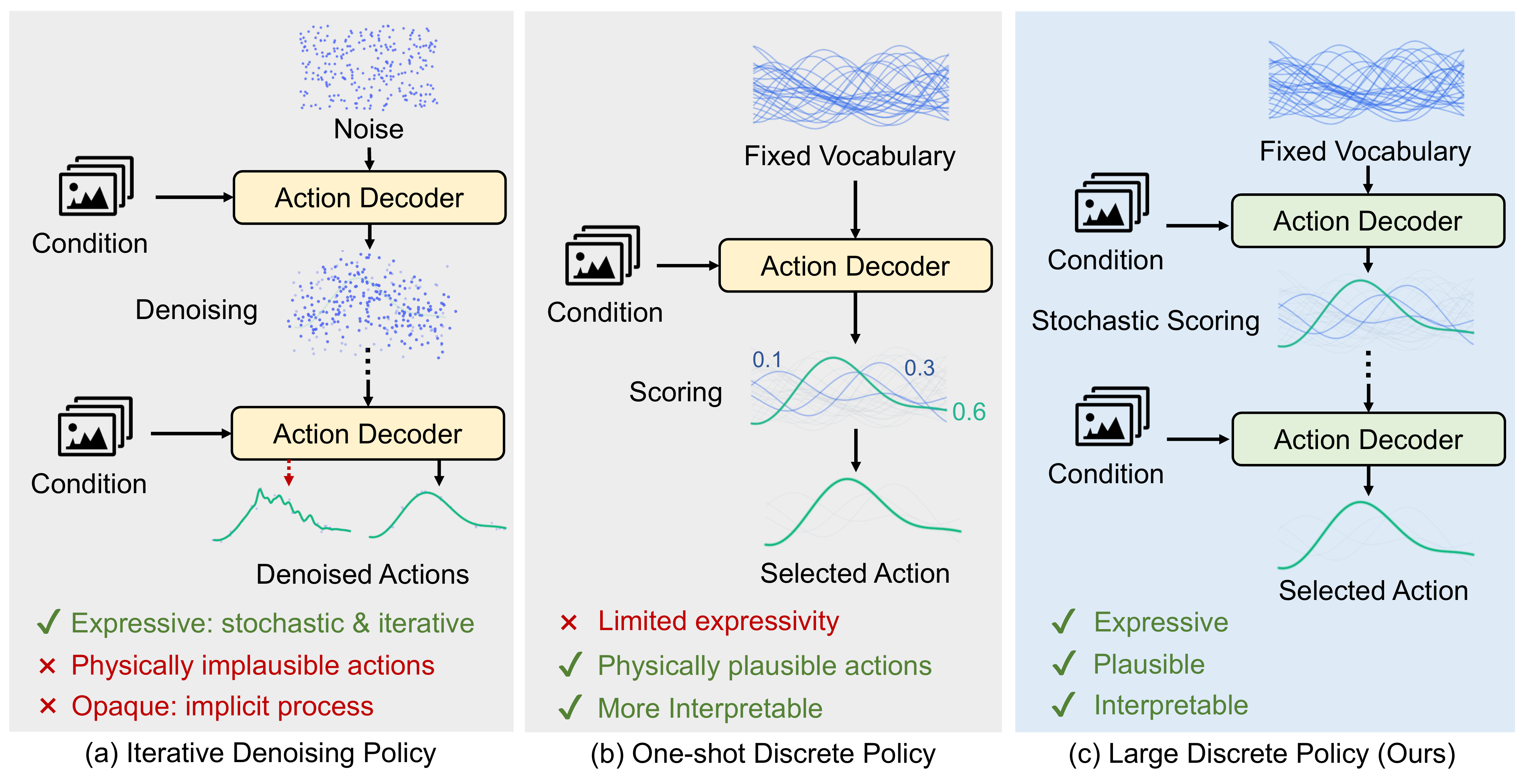}

    \caption{\textbf{Paradigms for Behavior Modeling Policies.}
    (a) Iterative denoising policies generate actions through stochastic iterative refinement, offering strong expressivity but remaining an implicit and less interpretable generation process.
    (b) One-shot discrete policies explicitly select from a fixed action vocabulary, but are limited by single-pass scoring.
    (c) Our approach combines the strengths of both paradigms by performing stochastic iterative scoring over a discrete action vocabulary, enabling expressive, plausible, and interpretable behavior generation.
    }

    \label{fig:teaser}
\end{figure*}

%% file: sections/2_related.tex
\vspace{-0.1in}
\section{Related Work}

\subsection{Iterative Denoising Policies for Behavior Modeling}

Iterative denoising policies have become a prominent paradigm for continuous behavior modeling. These methods formulate behavior modeling as an action generation problem. They are typically instantiated with diffusion models~\cite{ho2020denoising, song2021score, croitoru2023diffusion} or flow-matching models~\cite{lipman2023flow}. Diffusion Policy~\cite{chi2025diffusion} first demonstrated the effectiveness of diffusion-based action generation for robotic manipulation, showing strong capability in modeling multimodal action distributions from visual observations. Similar ideas have since been adopted in autonomous driving, including the DiffusionDrive series~\cite{liao2025diffusiondrive, zou2025diffusiondrivev2} for end-to-end planning with visual inputs, as well as planning-centric approaches~\cite{tan2025flow, zheng2025diffusionbased} that leverage privileged information (e.g. HD maps and 3D bounding boxes). More recently, the $\pi$ series~\cite{black2024pi_0, intelligence2025pi_} integrates a flow-based policy with pretrained Vision-Language Models, demonstrating promising open-world generalization. GR00T-N1~\cite{bjorck2025gr00t} similarly adopts diffusion-based action generation within a large-scale robot foundation model. While these approaches achieve strong empirical performance, their generation process remains largely implicit: actions are produced through repeated refinement of noisy samples, making it difficult to inspect why certain behaviors are considered and how the final behavior is generated.

\subsection{Discrete Policies for Behavior Modeling}

Discrete policies provide an alternative to continuous action generation by representing behaviors through a finite set of action candidates. Early work such as Behavior Transformers~\cite{shafiullah2022behavior} uses K-Means to construct action cluster centers and predicts continuous offsets from the selected cluster to the target behavior. While this introduces a discrete structure, the final action remains continuous and the policy is therefore not fully discrete. In autonomous driving, a line of trajectory scoring methods~\cite{philion2020lift, chen2024vadv2, li2024hydra, li2025hydranext, li2025generalized, li2025ztrs, sun2026sparsedrivev2} adopts a more explicit discrete formulation: they construct a vocabulary of trajectory candidates, often represented as waypoint sequences or action chunks, score each candidate under the current scene context, and select the highest-scoring trajectory for execution. More recent hierarchical trajectory scorers~\cite{yao2025drivesuprim, yao2026had} improve expressivity by first pruning a large candidate set and then refining a smaller subset. However, these methods often rely on additional refinement modules or inject noise directly into action candidates~\cite{yao2026had}, which may corrupt physically plausible trajectories and blur the fully discrete formulation. 

In the Vision-Language-Action literature, another family of methods discretizes actions into language tokens and predicts them through autoregressive generation, including manipulation models such as RT-1~\cite{brohan2022rt}, RT-2~\cite{zitkovich2023rt}, and OpenVLA~\cite{kim2024openvla}, as well as autonomous driving models~\cite{zhou2025autovla, luo2025adathinkdrive, dang2026drivefine}. These approaches benefit from the token-generation interface of pretrained language models, but they do not explicitly model actions in the manner of trajectory scorers. Our work builds on the explicitness of fully discrete trajectory scoring, and extends it into a general behavior modeling framework with a large action vocabulary and stochastic iterative scoring. With these components, \hydra{} improves expressivity while preserving interpretability and physical plausibility.

%% file: sections/3_method.tex
\section{Method}
\input{figures/arch_}

\label{sec:method}
\subsection{Model Architecture}

In this section, we describe the model architecture and learning procedure of \hydra{}, as illustrated in Fig.~\ref{fig:arch}. We begin by introducing how \hydra{} constructs the fixed action vocabulary and performs scoring in a single forward pass.

\textbf{Vocabulary Generation and Scoring.} Following LSS~\cite{philion2020lift}, we formulate behavior generation as discrete scoring over an action vocabulary. Given an offline demonstration dataset, we first segment expert behaviors into fixed-horizon action chunks $\mathbf{a}_{1:H}\in\mathbb{R}^{H\times d}$, where $H$ is the prediction horizon and $d$ is the action dimension. We then apply K-Means clustering to construct a large discrete vocabulary $\mathcal{V}$ with $K$ action candidates following~\cite{shafiullah2022behavior, philion2020lift, li2024hydra}:
\begin{equation}
    \mathcal{V}=\{\mathbf{v}_i\}_{i=1}^{K}
    =\operatorname{K-Means}\left(\{\mathbf{a}^{(n)}_{1:H}\}_{n=1}^{N}\right),
\end{equation}
where each cluster center $\mathbf{v}_i$ corresponds to a physically plausible action chunk observed in the demonstration distribution. Given visual observations, and optionally language instructions in the VLA setting, an observation encoder produces condition tokens $\mathbf{C}$. Each action chunk $\mathbf{v}_i$ is mapped into an action token by an action tokenizer $\phi(\cdot)$, and serves as the query to a Transformer decoder~\cite{vaswani2017attention}. The decoder attends to the condition tokens and predicts scores for candidate action chunks:
\begin{equation}
    \mathbf{h}^{(t)}_i
    = \operatorname{Dec}_{\theta}\left(\phi(\mathbf{v}^{(t)}_i), \mathbf{C}\right),
    \qquad
    s^{(t)}_i = g_{\theta}\left(\mathbf{h}^{(t)}_i\right),
\end{equation}
where $\mathcal{V}^{(t)}=\{\mathbf{v}^{(t)}_i\}_{i=1}^{K^{(t)}}$ is the candidate set at the $t$-th scoring stage, and $g_{\theta}$ is an MLP scoring head.

\textbf{Iterative Scoring.} Instead of scoring the vocabulary only once, \hydra{} performs coarse-to-fine iterative scoring. The first stage evaluates the full vocabulary $\mathcal{V}^{(1)}=\mathcal{V}$. Subsequent stages keep only the top-ranked candidates from the previous stage,
\begin{equation}
    \mathcal{I}^{(t+1)}
    = \operatorname{TopK}\left(\mathbf{s}^{(t)}, K^{(t+1)}\right),
    \qquad
    \mathcal{V}^{(t+1)}
    = \{\mathbf{v}^{(t)}_i \mid i\in\mathcal{I}^{(t+1)}\},
\end{equation}
where $\mathcal{I}$ denotes the action indices.
\hydra{} then re-scores this smaller candidate set with the shared action decoder ($\operatorname{Dec}_{\theta}$ and $g_{\theta}$). After repeating this process iteratively for $T$ stages, the policy outputs the highest-scoring action chunk:
\begin{equation}
    \hat{\mathbf{a}}_{1:H}
    =
    \mathbf{v}^{(T)}_{\arg\max_i s^{(T)}_i}.
\end{equation}
This design preserves a fully discrete decision process while allowing the model to progressively prune the vocabulary to the most promising regions of the action space. 

\textbf{Model Learning.} For imitation learning, we follow the soft target formulation used in Hydra-MDP~\cite{li2024hydra}. Given a ground-truth action chunk $\mathbf{a}^{*}_{1:H}$, each candidate receives a distance-based target score, which is converted into a soft-label distribution:
\begin{equation}
    q^{(t)}_i
    =
    \frac{
    \exp\left(-\|\mathbf{v}^{(t)}_i-\mathbf{a}^{*}_{1:H}\|^2_2/\sigma\right)
    }{
    \sum_j \exp\left(-\|\mathbf{v}^{(t)}_j-\mathbf{a}^{*}_{1:H}\|^2_2/\sigma\right)
    },
\end{equation}
where $\sigma$ is a hyperparameter that controls the sharpness of the distribution. The imitation objective at each stage is then formulated as the cross-entropy between the predicted logit scores $s^{(t)}$ and the soft label $q^{(t)}$:
\begin{equation}
    \mathcal{L}^{(t)}_{\mathrm{imi}}
    =
    -\sum_i q^{(t)}_i
    \log
    \frac{\exp(s^{(t)}_i)}{\sum_j\exp(s^{(t)}_j)}.
\end{equation}
For robotic manipulation, we use only this imitation objective, summed over all scoring stages. For autonomous driving, following trajectory scorers~\cite{li2024hydra, li2025hydra, li2025ztrs, sun2026sparsedrivev2}, we additionally attach multiple scoring heads to predict open-loop driving objectives $\mathcal{M}$~\cite{dauner2024navsim, cao2025pseudo}, including collision avoidance, drivable-area compliance, and related driving criteria. The driving loss is
\begin{equation}
    \mathcal{L}_{\mathrm{drive}}
    =
    \sum_{t=1}^{T}
    \left(
    \lambda_{\mathrm{imi}}\mathcal{L}^{(t)}_{\mathrm{imi}}
    +
    \sum_{m\in\mathcal{M}}
    \lambda_m \mathcal{L}^{(t)}_m
    \right),
\end{equation}
where $\mathcal{L}^{(t)}_m$ is implemented as binary cross-entropy between the predicted score and the ground-truth label. In this setting, the top-k selection in the iterative scoring process is based on a weighted average of multiple predicted scores, following Hydra-MDP~\cite{li2024hydra}.

\textbf{Stochastic Scoring.} During training, we introduce stochasticity in the pruning process rather than perturbing the action chunks themselves. Specifically, before top-k selection at intermediate stages, we perturb the original scores with noise sampled from a Gumbel distribution~\cite{jang2016categorical}:
\begin{equation}
    \tilde{s}^{(t)}_i = s^{(t)}_i + \tau \epsilon_i,
    \qquad
    \epsilon_i \sim \operatorname{Gumbel}(0,1).
\end{equation}
The next candidate set is selected by $\operatorname{TopK}(\tilde{s}^{(t)}, K^{(t+1)})$ during training.
For inference, we empirically find that using deterministic scores from the model yields the best performance in driving, while using noise-perturbed scores for selection improves online exploration in manipulation tasks, leading to improved manipulation performance.
This score-space perturbation encourages exploration among plausible action candidates without corrupting the discrete vocabulary or producing physically implausible action candidates.

\input{figures/vla_}
\subsection{Vision-Language-Action (VLA) Architecture}
Fig.~\ref{fig:vla} shows three variants for extending \hydra{} to the VLA setting: \begin{itemize}[
    label=\textbullet,
    topsep=0.2\baselineskip,
    itemsep=0.2\baselineskip,
    parsep=0pt,
    leftmargin=*
]
    \item \textbf{Action Group Token.} This variant follows the $\pi$-style joint-attention design~\cite{black2024pi_0}: we tokenize the discrete action vocabulary into action group tokens, concatenate them with vision-language tokens from the VLM, and allow joint attention between action and vision-language representations. The output action-token features are then decoded into vocabulary scores through lightweight linear scoring heads, preserving the iterative scoring and pruning procedure in Sec.~\ref{sec:method}.
    \item \textbf{Implicit Action Token.} The second variant removes direct access to the action vocabulary from the VLM. Instead, the model receives a small number of learned implicit action tokens, which attend to vision-language tokens and are linearly decoded into scores over the current candidate set. This design maintains a small number of tokens and tests whether the VLA can learn action scoring without explicitly observing the vocabulary.
    \item \textbf{The Two-System Approach.} The final variant follows the design of GR00T-N1~\cite{bjorck2025gr00t}: a frozen VLM first encodes visual observations and language instructions into vision-language tokens, which are then projected to the action-model dimension and used as conditioning tokens for a separate action decoder. The action model performs the same action selection process as described above.
\end{itemize}   

%% file: figures/arch_.tex
\begin{figure*}[tp]
    \centering
    \includegraphics[width=\linewidth]{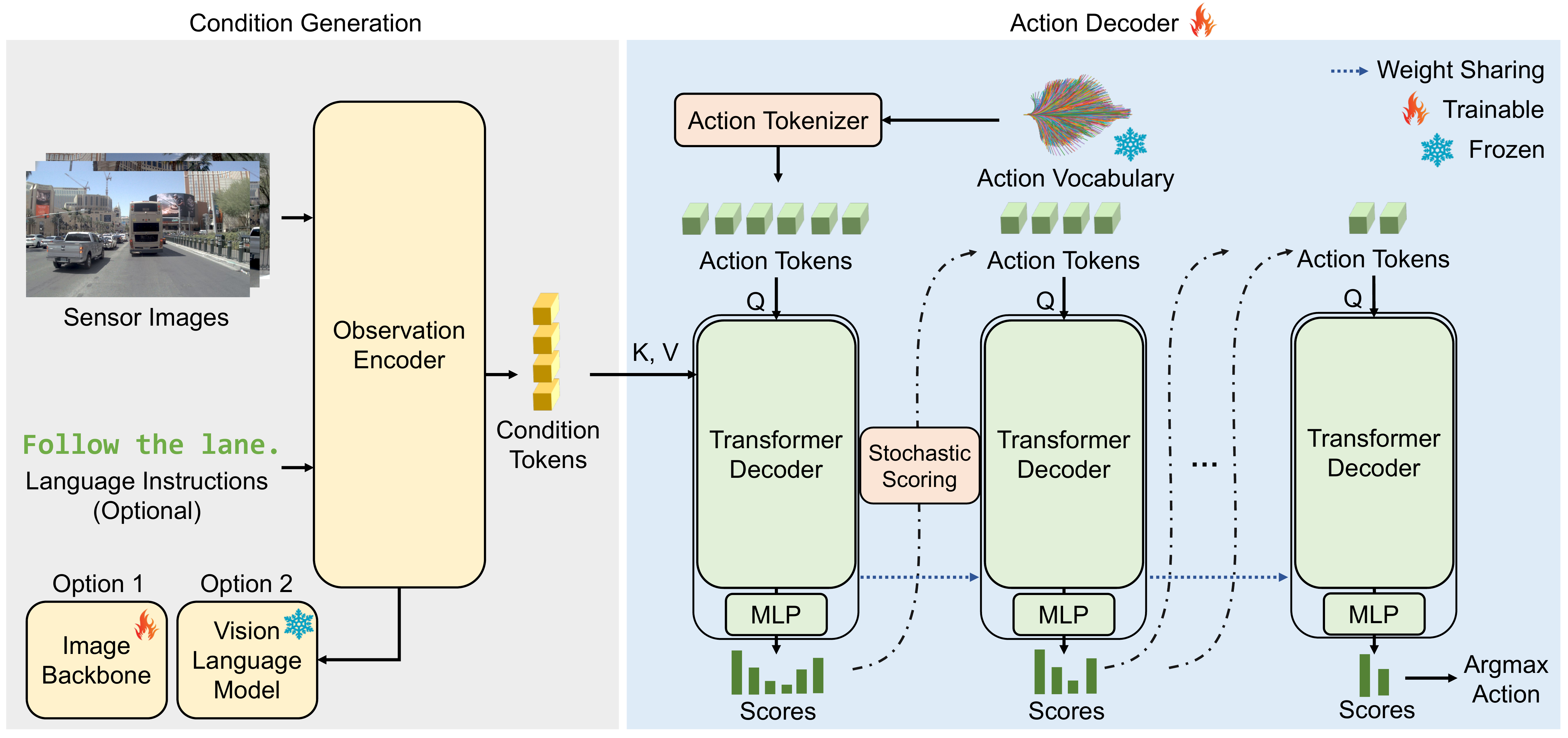}
    \vspace{-0.2in}

    \caption{\textbf{The Model Architecture of \hydra.} \hydra{} first tokenizes a fixed action vocabulary into action tokens, which attend to visual observations and optional language instructions in a Transformer decoder. The model then scores all candidates with an MLP, stochastically prunes the vocabulary, and iteratively re-scores the remaining candidates until a single action is selected.}
    \vspace{-0.15in}

    \label{fig:arch}
\end{figure*}

%% file: figures/vla_.tex
\begin{figure*}[!t]
    \centering
    \includegraphics[width=\linewidth]{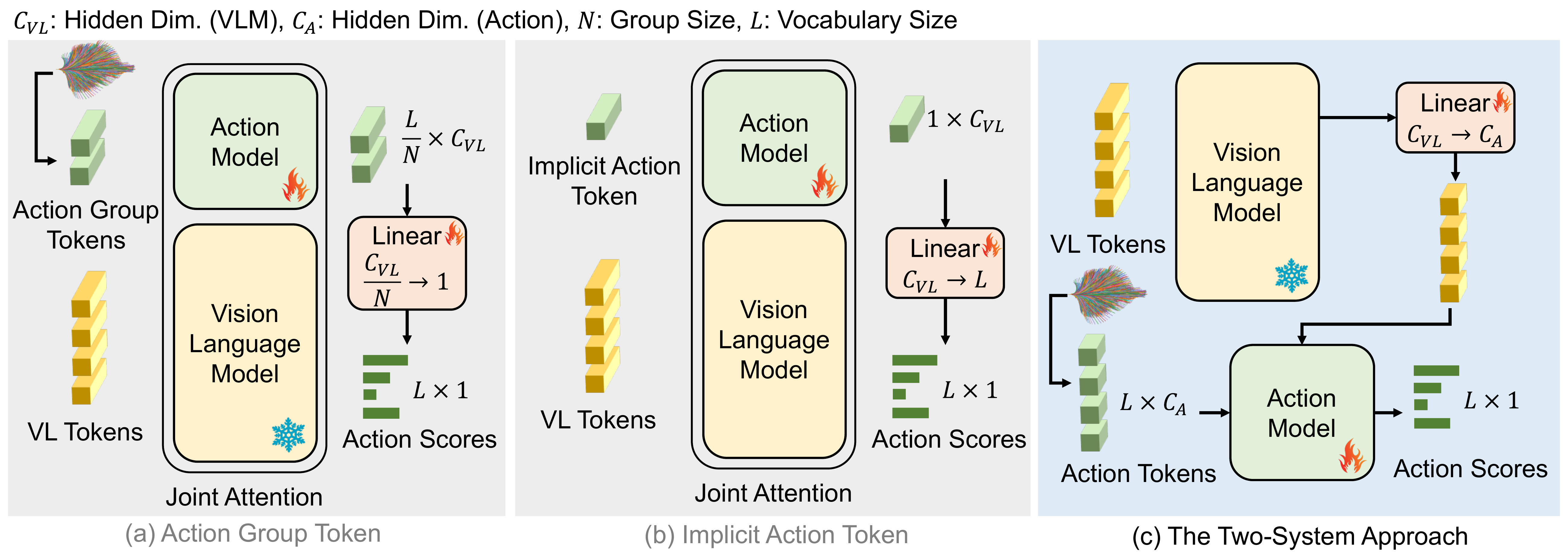}
    \vspace{-0.2in}

    \caption{\textbf{Vision-Language-Action (VLA) Architecture Variants.}}
    \vspace{-0.2in}

    \label{fig:vla}
\end{figure*}

%% file: sections/4_exp.tex
\input{tables/navsim}
\input{tables/navsim_vla}
\vspace{-0.1in}
\section{Experiments}
\subsection{Implementation Details}

We train our autonomous driving models on the NAVSIM Navtrain split~\cite{dauner2024navsim}, which contains 103K scenes. For standard end-to-end planning models, we train with 24 NVIDIA A100 GPUs for 20 epochs using Adam~\cite{kingma2014adam} with a learning rate of $2\times10^{-4}$ and a per-GPU batch size of 12. For VLAs, we train with 8 NVIDIA A100 GPUs for 8 epochs using Adam with a learning rate of $7.5\times10^{-5}$ and a per-GPU batch size of 8. The action vocabulary contains 16,384 trajectories, each spanning 4 seconds at 10 Hz. Unless specified, the vocabulary is pruned from 16,384 to 512 and then to 16 candidates. 
On HUGSIM~\cite{zhou2024hugsim}, we perform zero-shot closed-loop inference following~\cite{li2025ztrs}. For the controller variant, we truncate the predicted trajectory to 3 seconds and fix the heading conversion~\cite{kirby2026drivor}.

For robotic manipulation, we largely follow the training and evaluation protocol of Diffusion Policy~\cite{chi2025diffusion}, which is also our major baseline. We use action chunks with a horizon length of 16, 2 observation steps, and execute 8 steps per prediction. Models are trained with AdamW~\cite{loshchilov2017decoupled} using a learning rate of $1\times10^{-4}$ and task-dependent batch sizes following the Diffusion Policy setup. The vocabulary size is 8,192 for PushT and Can, 16,384 for ToolHang due to its higher dexterity requirement, and 6,666 for Lift, corresponding to the number of available training trajectories. The remaining pruning stages are consistent with autonomous driving.

\vspace{-0.1in}
\subsection{Benchmarks and Metrics}
\textbf{Autonomous driving.}
We evaluate autonomous driving on NAVSIM~\cite{dauner2024navsim, cao2025pseudo} and HUGSIM~\cite{zhou2024hugsim}. NAVSIM contains 103K training scenes in Navtrain and 12k evaluation scenes in Navtest. For standard discrete end-to-end models, we report Navtest v2 with the Extended Predictive Driver Model Score (EPDMS)~\cite{dauner2024navsim, cao2025pseudo, li2025hydra}, which aggregates No-at-fault Collisions (NC), Drivable Area Compliance (DAC), Driving Direction Compliance (DDC), Traffic Light Compliance (TLC), Ego Progress (EP), Time-to-Collision (TTC), Lane Keeping (LK), History Comfort (HC), and Extended Comfort (EC). For VLA models, we follow recent VLA works and report Navtest v1 with PDMS. On the other hand, HUGSIM evaluates zero-shot closed-loop performance in photo-realistic scenarios built with 3D Gaussian Splatting~\cite{kerbl20233d}. It contains 345 scenarios from KITTI-360~\cite{liao2022kitti}, Waymo~\cite{sun2020scalability}, nuScenes~\cite{caesar2020nuscenes}, and PandaSet~\cite{xiao2021pandaset}. Easy scenarios involve regular driving, Medium scenarios introduce inserted vehicles, and Hard/Extreme scenarios include aggressive vehicles. We report Route Completion (RC) and HD-Score, which jointly measure progress and closed-loop safety.

\textbf{Robotic manipulation.}
We evaluate visual-based robotic manipulation following Diffusion Policy~\cite{chi2025diffusion}. The benchmark includes three Robomimic tasks~\cite{Mandlekaretal2022}, Lift, Can, and ToolHang, where the policy controls a 6-DoF end-effector and gripper from visual observations. We also evaluate on PushT, a 2-DoF task that requires pushing a T-shaped block to a target pose.

\input{tables/hugsim}
\input{tables/robotics}
\vspace{-0.1in}
\subsection{Main Results}

\textbf{NAVSIM, End-to-end Planning.}
Tab.~\ref{tab:navsim} shows that \hydra{} achieves the best EPDMS across all evaluated backbones, reaching 84.7 with ResNet34~\cite{he2016deep}, 88.1 with ViT-L~\cite{dosovitskiy2020image}, and 87.8 with V2-99~\cite{lee2019energy}. The gains are consistent across CNN- and Transformer-based architectures, indicating that \hydra{} is not tied to a particular visual encoder. Instead, its advantage comes from the action decoder: by scoring a large vocabulary iteratively and stochastically, \hydra{} outperforms both discrete trajectory scorers~\cite{li2025hydra, li2025generalized, li2025ztrs} and diffusion-based policies~\cite{liao2025diffusiondrive, chi2025diffusion} by a significant margin.

\textbf{NAVSIM, Vision-Language-Action Models.}
Tab.~\ref{tab:navsim_vla} further shows that the action modeling strategy of \hydra{} transfers effectively to vision-language representations. Despite using a compact Qwen3-VL-2B~\cite{bai2025qwen3} backbone, \hydra{}-VLA achieves the best overall PDMS of 92.1, outperforming existing VLA baselines. This result highlights an important property of \hydra{}: its scoring mechanism is robust across substantially different architectures, from perception-only planning models to Vision-Language-Action models. The improvement suggests that \hydra{} provides a transferable action modeling interface, rather than an architecture-specific enhancement.

\textbf{HUGSIM, Closed-loop Driving.}
Tab.~\ref{tab:hugsim} evaluates zero-shot closed-loop driving on HUGSIM. \hydra{} achieves strong closed-loop performance without fine-tuning on HUGSIM, and the controller variant obtains the best overall RC of 44.7 and HD-Score of 36.7. Compared with prior methods, the gains are especially clear on Easy and Medium scenarios, suggesting that \hydra{} transfers well from open-loop training to closed-loop execution when the domain gap is moderate. However, we observe a performance drop in Extreme scenarios after controller tuning, suggesting that these cases remain highly challenging despite the advancements in action modeling.

\textbf{Visual-based Robotic Manipulation.}
Our results demonstrate that \hydra{} can transfer across task domains from autonomous driving to robotic manipulation. As shown in Tab.~\ref{tab:robotics_control}, \hydra{} achieves success rates of 0.99 on Lift and 0.94 on Can, remaining competitive with the Transformer-based Diffusion Policy~\cite{chi2025diffusion}. More importantly, it outperforms Diffusion Policy on ToolHang and PushT, reaching 0.52 and 0.76 success rates, respectively. These results show that stochastic discrete scoring is not limited to trajectory planning, but also applies to high-dimensional manipulation.

\input{tables/abl_arch}

\input{tables/abl_vla_robotics}

\vspace{-0.1in}
\subsection{Ablation Studies}

\textbf{Key Components.}
Tab.~\ref{table:ablation_arch} ablates the main components of \hydra{}. Starting from Hydra-MDP-$\mathcal{V}_{16384}$, iterative scoring improves EPDMS from 86.2 to 86.7, showing the benefit of re-scoring a pruned candidate set. We further introduce an architectural improvement by replacing the MLP action tokenizer used in prior trajectory scorers~\cite{li2024hydra, li2025generalized} with a 1D convolutional tokenizer, which facilitates action-wise temporal modeling. We also adopt a cross-attention-only Transformer decoder, where action tokens no longer perform self-attention over the full vocabulary. This improves EPDMS to 87.1 while remaining efficient for large vocabularies. Adding stochastic scoring achieves the best EPDMS of 87.8. Compared with injecting noise into top-ranked actions, which reaches 87.2, stochastic scoring performs better while keeping the action vocabulary physically plausible.

\textbf{VLA Architectures.}
The ablation in Tab.~\ref{tab:abl_vla} shows that stochastic iterative scoring consistently improves all three VLA designs. The Two-System approach performs best, suggesting that separating vision-language understanding from low-level action scoring~\cite{gr00tn1_2025} is more effective than directly folding action scoring into the VLM. The Action Group Token variant faces the challenge of too many action tokens during joint attention, as the original vocabulary can contain up to 16K candidates and must be downsampled into fewer groups, (e.g. 2,048 group tokens). In contrast, the Implicit Action Token variant does not access the vocabulary directly, which reduces token cost but loses fine-grained geometric information for candidate selection.

\textbf{Robotic Manipulation.}
Tab.~\ref{tab:robotics_ablation} shows that the same design choices transfer to manipulation. Iterative pruning improves both ToolHang and PushT over the one-shot discrete policy, while stochastic scoring gives the best success rates, improving ToolHang from 0.54 to 0.60 and PushT from 0.67 to 0.78. This confirms that stochastic score perturbation is useful beyond autonomous driving.

\textbf{Width and Depth.}
Fig.~\ref{fig:abl_width_depth} studies the number of scoring stages and vocabulary sizes. Performance improves from one to three stages, reaching 87.8 EPDMS, but slightly drops with four stages, indicating that additional pruning stages bring diminishing returns. For vocabulary width, the best setting uses 512 intermediate candidates and 16 final candidates. Smaller candidate sets lose useful actions too early, while oversized sets make final scoring more ambiguous.

\input{figures/vis_}
\input{figures/abl_width_depth}
\vspace{-0.1in}
\subsection{Qualitative Results}

Fig.~\ref{fig:vis} visualizes the iterative scoring process of \hydra{} on Navtest. At early stages, the remaining vocabulary covers diverse plausible futures, while later stages progressively concentrate on a smaller set of scene-consistent trajectories. (a): \hydra{} anticipates nearby traffic participants and prunes candidates toward a socially compliant trajectory. (b): the model preserves aggressive candidates in early stages but shifts towards a safer selection in response to the environmental uncertainty. (c): \hydra{} reasons about potentially occluded, distant agents and selects a conservative trajectory. 
\vspace{-0.1in}

%% file: tables/navsim.tex
\begin{table}[!t]
    \centering

\caption{\textbf{Performance on the Navtest v2 Benchmark, End-to-end Planning Models.}}
\vspace{-0.05in}
    \small
    \resizebox{\textwidth}{!}{
    \begin{tabular}{c|c|c|ccccccccc|c}
        \toprule
        Backbone & Method & Behavior Modeling  & NC $\uparrow$ & DAC $\uparrow$ & DDC $\uparrow$ & TLC $\uparrow$ & EP $\uparrow$ & TTC $\uparrow$ & LK $\uparrow$ & HC $\uparrow$ & EC $\uparrow$ & EPDMS $\uparrow$ \\
        \midrule
        - & Human Agent & -  & 100 & 100 & 99.8 & 100 & 87.4 & 100 & 100 & 98.1 & 90.1 & 90.3 \\
        \midrule
        \midrule
        - & Ego Status MLP & Regression & 93.1 & 77.9 & 92.7 & 99.6 & 86.0 & 91.5 & 89.4 & 98.3 & 85.4 & 64.0 \\
        \midrule

        \multirow{6}{*}{ResNet34} & Transfuser~\citep{chitta2022transfuser} & Regression & 96.9 & 89.9 & 97.8 & 99.7 & 87.1 & 95.4 & 92.7 & \textbf{98.3} & 87.2 & 76.7 \\
        & HydraMDP++~\citep{li2025hydra} & Discrete & 97.2 & 97.5 & 99.4 & 99.6 & 83.1 & 96.5 & 94.4 & 98.2 & 70.9 & 81.4 \\
        & DriveSuprim~\citep{yao2025drivesuprim} & Discrete & 97.5 & 96.5 & 99.4 & 99.6 & 88.4 & 96.6 & 95.5 & \textbf{98.3} & 77.0 & 83.1 \\
        &  ZTRS~\citep{li2025ztrs} & Discrete &  97.9 &  \textbf{98.0} &  \textbf{99.7} &  \textbf{99.8} &  80.1 &  96.9 &  94.9 &  98.1 &  79.0 &  83.2 \\
        &  DiffusionDrive~\citep{liao2025diffusiondrive} & Diffusion &  \textbf{98.4} &  95.5 &  99.5 &  \textbf{99.8} &  87.5 &  \textbf{97.5} &  \textbf{96.9} &  98.4 &  \textbf{87.7} &  84.2 \\
        
        \cmidrule{2-13}
        &  \hydra (Ours) & Discrete &  97.7 &  97.5 &  99.4 &  99.7 &  \textbf{88.5} &   97.0 &  96.6 &  \textbf{98.3} &  79.4 &  \textbf{84.7} \\
        \midrule

        \multirow{5}{*}{ViT-L} & HydraMDP++~\citep{li2025hydra} & Discrete & 98.5 & 98.5 & 99.5 & 99.7 & 87.4 & 97.9 & 95.8 & 98.2 & 75.7 & 85.6 \\
        & DriveSuprim~\citep{yao2025drivesuprim} & Discrete & 98.4 & 98.6 & 99.6 & 99.8 & \textbf{90.5} & 97.8 & 97.0 & \textbf{98.3} & 78.6 & 87.1 \\
        & GTRS-Dense~\citep{li2025generalized} & Discrete & \textbf{99.0} & 97.8 & 99.3 & \textbf{99.9} & 86.3 & \textbf{98.3} & 95.1 & \textbf{98.3} & 71.9 & 84.7 \\
        &  ZTRS~\citep{li2025ztrs}  & Discrete &  98.2 &  \textbf{99.1} &  \textbf{99.7} &  99.8 &  86.9 &  97.5 &  96.6 &  98.2 &  78.2 &  86.2 \\
        \cmidrule{2-13}

        &  \hydra (Ours)  & Discrete &  98.7 &  \textbf{99.1} &  \textbf{99.7} &  99.8 &  90.3 &   98.0 &  \textbf{97.9} &  98.2 &  \textbf{80.7} &  \textbf{88.1} \\
        \midrule
        \multirow{6}{*}{V2-99}       
        & Diffusion Policy~\citep{chi2025diffusion} & Diffusion &  96.8 & 92.8 &  98.7 &  99.7 &  86.6 &  95.9 &  94.9 & 98.2 &  \textbf{83.7} &  79.1 \\ 
        &  HydraMDP++~\citep{li2025hydra} & Discrete & \textbf{98.4} & 98.0 & 99.4 & 99.8 & 87.5 & 97.7 & 95.3 & \textbf{98.3} & 77.4 & 85.1 \\
        & DriveSuprim~\citep{yao2025drivesuprim} & Discrete & 97.8 & 97.9 & 99.5 & \textbf{99.9} & 90.6 & 97.1 & 96.6 & \textbf{98.3} & 77.9 & 86.0 \\
        & GTRS-Dense~\citep{li2025generalized} & Discrete & 97.6 & 98.5 & 99.5 & \textbf{99.9} & 89.5 & 97.2 & 96.8 & 97.2 & 57.2 & 84.0 \\
        &  ZTRS~\citep{li2025ztrs}  & Discrete &  97.8 &  \textbf{99.4} &  \textbf{99.8} &  99.8 &  84.1 &  97.0 &  96.2 &  98.2 &  77.2 &  85.3 \\

        \cmidrule{2-13}
        &  \hydra (Ours)  & Discrete &  98.3 &  98.9 &  99.7 &  99.8 &  \textbf{91.0} &   \textbf{97.9} &  \textbf{97.5} &  98.1 &  79.1 &  \textbf{87.8} \\
        
        \bottomrule
    \end{tabular}
    }
    \label{tab:navsim}
    
\end{table}

%% file: tables/navsim_vla.tex
\begin{table}[!t]
\centering
\vspace{-0.1in}
\caption{\textbf{Performance on the Navtest v1 Benchmark, Vision-Language-Action Models.}}
\vspace{-0.05in}

\resizebox{\textwidth}{!}{
\begin{tabular}{l|c|c|ccccc|c}
\toprule
Method & VLM & Behavior Modeling & NC$\uparrow$ & DAC$\uparrow$ & TTC$\uparrow$ & C$\uparrow$ & EP$\uparrow$ & PDMS$\uparrow$ \\ \midrule
Human & -- & -- & 100 & 100 & 100 & 99.9 & 87.5 & 94.8 \\ \midrule \midrule
AutoVLA~\cite{zhou2025autovla} & Qwen2.5-VL-3B~\cite{Qwen2.5-VL} & Token Generation & 98.4 & 95.6 & \textbf{98.0} & 99.9 & 81.9 & 89.1 \\
ReCogDrive~\cite{li2025recogdrive} & InternVL3-8B~\cite{zhu2025internvl3} & Diffusion & 98.2 & 97.8 & 95.2 & 99.8 & 83.5 & 89.6 \\
DriveVLA-W0~\cite{li2025drivevla} & Emu3-8B~\cite{wang2024emu3} & Discrete  & 98.7 & \textbf{99.1} & 95.3 & 99.3 & 83.3 & 90.2 \\
AdaThinkDrive~\cite{luo2025adathinkdrive} & InternVL3-8B~\cite{zhu2025internvl3} & Token Generation & 98.4 & 97.8 & 95.2 & \textbf{100} & 84.4 & 90.3 \\
SpanVLA~\cite{zhou2026spanvla} & Qwen2.5-VL-3B~\cite{Qwen2.5-VL} & Flow Matching & \textbf{99.1} & 97.1 & 95.2 & \textbf{100} & 86.3 & 90.3 \\

DriveFine~\cite{dang2026drivefine} & Lavida-8B~\cite{li2025lavida} & Token Generation & 98.8 & 98.6 & 96.2 & \textbf{100} & 86.9 & 91.8 \\ 

\midrule

\hydra-VLA (Ours) & Qwen3-VL-2B~\cite{bai2025qwen3} & Discrete & 99.0 & 98.1 & 98.6 & 98.2 & \textbf{88.8} & \textbf{92.1} \\
\bottomrule
\end{tabular}
}
\label{tab:navsim_vla}
\vspace{-0.2in}
\end{table}

%% file: tables/hugsim.tex
\begin{table}[!t]
\centering

\caption{\textbf{Zero-shot Performance on the HUGSIM Benchmark.} UniAD* and VAD* are trained on nuScenes~\cite{caesar2020nuscenes}, which partially overlap with the evaluation scenarios. Other methods are evaluated with zero-shot transfer. $\dagger$ The controller is tuned for closed-loop driving.}
\vspace{-0.05in}
\resizebox{\textwidth}{!}{
\begin{tabular}{l|cccccccc|cc}
\toprule
\multirow{2}{*}[-2pt]{Method} 
& \multicolumn{2}{c}{Easy} 
& \multicolumn{2}{c}{Medium} 
& \multicolumn{2}{c}{Hard} 
& \multicolumn{2}{c|}{Extreme} 
& \multicolumn{2}{c}{Overall} \\
\cmidrule{2-11}
 & RC $\uparrow$ & HD-Score $\uparrow$ & RC $\uparrow$ & HD-Score $\uparrow$ & RC $\uparrow$ & HD-Score $\uparrow$ & RC $\uparrow$ & HD-Score $\uparrow$ & RC $\uparrow$ & HD-Score $\uparrow$ \\
\midrule

VAD*~\citep{jiang2023vad}
& 38.7 & 24.3 & 27.0  & 9.9  & 25.5  & 10.4  & 23.0  & 8.2  &  27.9 & 12.3  \\

UniAD*~\citep{hu2023planning}
& 58.6 & 48.7 &  41.2 &  29.5 & \textbf{40.4}  & \textbf{27.3}  & 26.0  &  14.3 & 40.6  &  28.9 \\

LTF~\citep{chitta2022transfuser} 
& 68.4 & 52.8 & 40.7  & 24.6  & 36.9  &  19.8 & 25.5  &  8.1 &  41.4 &  24.8 \\

Diffusion Policy~\citep{chi2025diffusion} 
& 72.4 &  60.8 &  30.5 &  15.2 & 27.4 & 13.4  &  28.0 & 14.9  &  36.9 &  23.1 \\

ZTRS~\citep{li2025ztrs}
& 74.8  &  66.9 &  46.1 &  37.9 & 31.6 & 21.1  &  20.0 & 9.8  &  42.0 &  32.9 \\

GTRS-Dense~\citep{li2025generalized} 
& 62.0 & 57.8  & 40.5 & 35.2 & 23.9  &  16.2 & 13.7  & 5.8  & 34.5  & 28.2 \\

\midrule

\hydra (Ours) 
& 80.5  &  71.1 &  38.7 &  22.8 & 30.7 & 16.1  &  \textbf{31.7} & \textbf{16.0}  &  43.0 &  28.6 \\

\hydra (Ours) $\dagger$
& \textbf{85.5}  &  \textbf{82.0} &  \textbf{50.3} &  \textbf{40.7} & 29.0 & 20.6  &  19.4 & 10.6  &  \textbf{44.7} &  \textbf{36.7} \\

\bottomrule
\end{tabular}
}

\label{tab:hugsim}
\end{table}

%% file: tables/robotics.tex
\begin{table}[t]
\centering
\vspace{-0.1in}
\caption{\textbf{Performance on Visual-based Robotic Manipulation}. Following Diffusion Policy~\cite{chi2025diffusion}, we report averaged success rates over the last 10 checkpoints. We compare against the Transformer-based Diffusion Policy, as both methods adopt a Transformer Decoder for behavior modeling.}
\vspace{-0.05in}

\label{tab:robotics_control}
\resizebox{0.7\linewidth}{!}{
\begin{tabular}{l|c|ccc|c}

\toprule
 & & IBC~\cite{florence2021implicit} & LSTM-GMM~\cite{Mandlekaretal2022} & Diffusion Policy~\cite{chi2025diffusion} & \hydra{} (Ours) \\ 
\cmidrule(lr){3-5} \cmidrule(lr){6-6}
\multirow{-2}{*}{Task} & \multirow{-2}{*}{DoF} & \multicolumn{3}{c|}{Continuous} & Discrete \\
\midrule
Lift  & 6+1 &
0.73 & 0.96 & \textbf{1.00} & 0.99 \\

Can & 6+1 & 
0.01 & 0.88 & \textbf{0.98} & 0.94 \\

ToolHang & 6+1 & 
0.00 & 0.49 & 0.47 & \textbf{0.52} \\

PushT & 2 & 
0.64 & 0.54 & 0.66 & \textbf{0.76} \\

\bottomrule
\end{tabular}
}
\vspace{-0.2in}
\end{table}

%% file: tables/abl_arch.tex
\begin{table}[t!]
\scriptsize
\centering
\caption{\textbf{Ablation Study on Key Components. }
}
\vspace{-0.05in}
\resizebox{\textwidth}{!}{
\begin{tabular}{l| c c c c c c c c c | c}

    \toprule
    Model Variants  

    & {NC} $\uparrow$
    & {DAC} $\uparrow$
    & {DDC} $\uparrow$
    & {TLC} $\uparrow$
    & {EP} $\uparrow$
    & {TTC} $\uparrow$
    & {LK} $\uparrow$
    & {HC} $\uparrow$
    & {EC} $\uparrow$
    & {EPDMS} $\uparrow$ \\
    \midrule
    
    Hydra-MDP-$\mathcal{V}_{16384}$~\cite{li2024hydra} & \textbf{98.5} & 98.7 & 98.9 & \textbf{99.9} & 88.5 & \textbf{98.2} & 97.0 & \textbf{98.3} & \textbf{80.5} & 86.2 \\

    \quad + Iterative Scoring & 97.9 & 98.3 & 99.5 & 99.8 & 90.6 & 97.3 & \textbf{98.1} & 98.1 &  77.1 & 86.7 \\

    \quad + Improved Tokenizer \& Cross-Attention Decoder & 98.4 & 98.7 & \textbf{99.7} & 99.8 & 89.6 & 97.9 & 97.4 & \textbf{98.3} & 78.6 & 87.1 \\

    \quad + Stochastic Scoring & 98.3 & \textbf{98.9} & \textbf{99.7} & 99.8 & \textbf{91.0} & 97.9 & 97.5 & 98.1 &  79.1 & \textbf{87.8} \\
    \midrule
    
    Hydra-MDP-$\mathcal{V}_{16384}$~\cite{li2024hydra} & \textbf{98.5} & 98.7 & 98.9 & \textbf{99.9} & 88.5 & \textbf{98.2} & 97.0 & \textbf{98.3} & \textbf{80.5} & 86.2 \\
    \quad + Noise Injection on Top-8 Actions~\cite{yao2026had} & 98.6 & 98.7 & \textbf{99.7} & \textbf{99.9} & 88.7 & 98.2 & 97.0 & \textbf{98.3} & 80.2 & 87.2 \\

\bottomrule
\end{tabular}
}

\label{table:ablation_arch}
\end{table}

%% file: tables/abl_vla_robotics.tex
\begin{table}[!t]
\vspace{-0.1in}
\centering

\begin{minipage}[t]{0.55\textwidth}
\vspace{0pt}
\centering
\scriptsize
\caption{\textbf{Ablation Study on VLAs.}}
\label{tab:abl_vla}
\vspace{-0.05in}

\resizebox{1.0\linewidth}{!}{

\begin{tabular}{l|ccccc|c}
\toprule
Model Variants & NC$\uparrow$ & DAC$\uparrow$ & TTC$\uparrow$ & C$\uparrow$ & EP$\uparrow$ & PDMS$\uparrow$ \\
\midrule
\textbf{(a) Action Group Token} & 97.9 & 96.5 & 97.3 & 98.2 & 84.9 & 87.9 \\
+ Stochastic Iterative Scoring & 97.6 & 95.9 & 96.9 & 97.7 & \textbf{89.2} & 88.8 \\
\midrule
\textbf{(b) Implicit Action Token} & 97.6 & 97.3 & 96.9 & 97.0 & 88.3 & 89.4 \\
+ Stochastic Iterative Scoring & 98.7 & 96.7 & \textbf{99.3} & \textbf{98.3} & 88.8 & 90.6 \\
\midrule
\textbf{(c) The Two-System Approach} & \textbf{99.0} & 98.0 & 98.7 & 98.2 & 85.8 & 90.8 \\
+ Stochastic Iterative Scoring & \textbf{99.0} & \textbf{98.1} & 98.6 & 98.2 & 88.8 & \textbf{92.1} \\
\bottomrule
\end{tabular}

}

\end{minipage}\hfill%
\begin{minipage}[t]{0.44\textwidth}
\vspace{0pt}
\centering
\scriptsize
\caption{\textbf{Ablation Study on Manipulation.}}
\label{tab:robotics_ablation}
\vspace{0.0in}

\resizebox{\linewidth}{!}{
\renewcommand{\arraystretch}{1.1}
\begin{tabular}{l|cc}
\toprule
Model Variants at Epoch 50 & ToolHang  & PushT \\
\midrule
One-shot Discrete Policy & 0.54 & 0.67 \\
\quad + Iterative Scoring & 0.56 & 0.69 \\
\quad + Stochastic Scoring & \textbf{0.60} & \textbf{0.78} \\
\bottomrule
\end{tabular}
}
\end{minipage}
\vspace{-0.2in}
\end{table}

%% file: figures/vis_.tex
\begin{figure*}[!t]
    \centering
    \includegraphics[width=\linewidth]{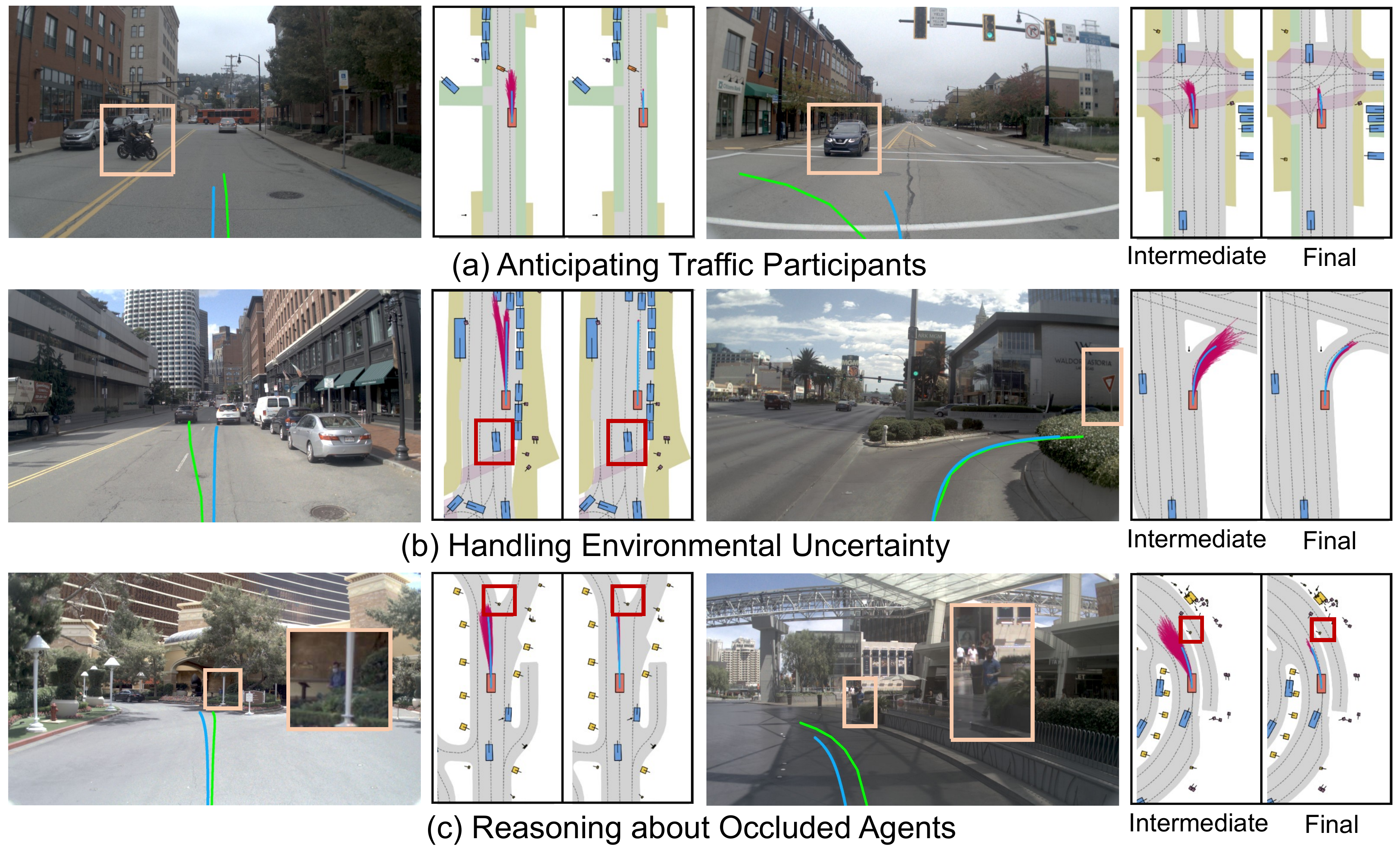}

    \vspace{-0.1in}
    \caption{\textbf{Visualizations on Navtest.} The figure shows the \textcolor{cyan}{selected trajectory}, the \textcolor{green}{human trajectory}, and the \textcolor{magenta}{remaining vocabulary} at the intermediate and final stages.}
    \vspace{-0.15in}

    \label{fig:vis}
\end{figure*}

%% file: figures/abl_width_depth.tex
\begin{figure*}[tp]
    \centering
    \includegraphics[width=\linewidth]{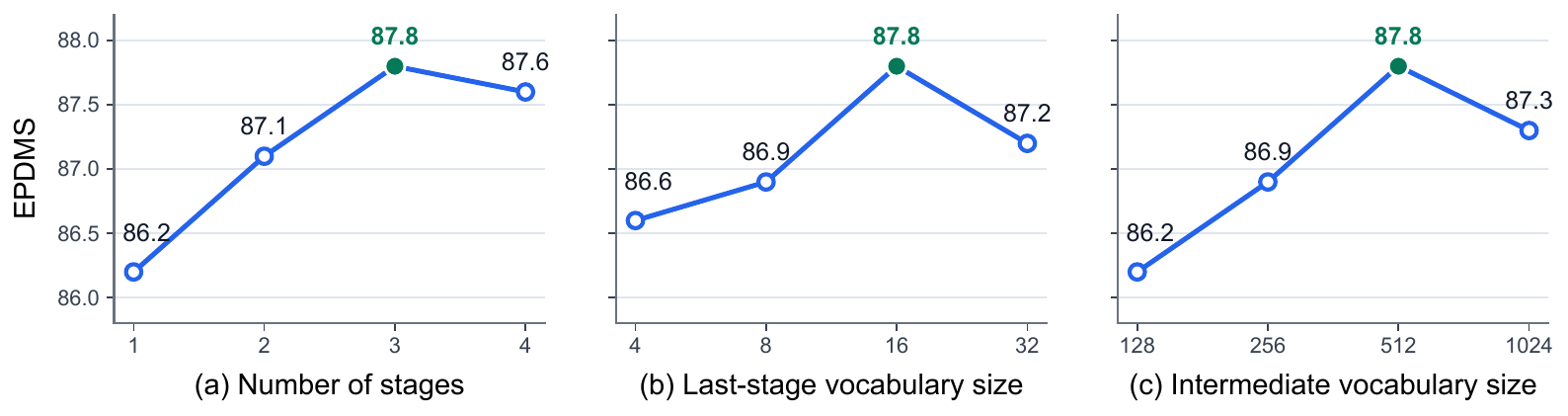}
\vspace{-0.2in}
    \caption{\textbf{Ablation Study on the Width and Depth of \hydra{}.} }
\vspace{-0.25in}
    \label{fig:abl_width_depth}
\end{figure*}

%% file: sections/5_conclusion.tex
\section{Conclusion}
We present the Large Discrete Policy (\hydra{}), a fully discrete behavior modeling framework that improves the expressivity of discrete policies without compromising their explicit and physically plausible formulation. By constructing a large fixed action vocabulary and introducing iterative scoring with stochastic score perturbation, \hydra{} progressively compares and prunes plausible action candidates while keeping the decision process interpretable. Across end-to-end planning, closed-loop driving, robotic manipulation, and vision-language-action settings, \hydra{} consistently demonstrates strong performance against both discrete trajectory scorers and iterative denoising policies. These results suggest that \hydra{} provides a practical and interpretable alternative for behavior modeling.

\textbf{Limitations.} While LDiP shows promising performance in behavior modeling, it still falls behind Diffusion Policy on certain manipulation tasks. Further evaluation in real-world settings is also needed. The scope of this work is to study real-world planning for autonomous driving and simulation-based closed-loop performance in both autonomous driving and robotic manipulation.

%% file: sections/7_ack.tex
\section{Acknowledgements}
This work is supported by the National Natural Science Foundation of China (Grant No. 62427819) and the Science and Technology Commission of Shanghai Municipality (No. 24511103100).

%% file: sections/6_appendix.tex
\section{Technical appendices and supplementary material}

Here we provide the technical appendices and supplementary material to complement the main paper. These materials include additional qualitative results and extended analyses of our approach.

\subsection{Additional Qualitative Results (Autonomous Driving)}

\input{figures/viz_supp_av_}

Fig.~\ref{fig:viz_supp_av} provides supplemental examples showing that our policy generates stable actions across diverse urban scenes. In (a) and (b), the model follows curved and straight lanes smoothly while remaining aligned with the intended drivable corridor, even in visually complex settings with parked vehicles, dense traffic, and distant traffic signs. Compared with the human trajectory, the predictions of the policy are more conservative, maintaining a larger safety buffer against traffic participants. The paired camera and Bird's-Eye-Views show that the predicted paths are not only visually plausible in image space, but also consistent with lane geometry and surrounding agents.

Fig.~\ref{fig:viz_supp_av} (c) further highlights interactive driving scenarios, where the ego vehicle must plan around nearby vehicles, pedestrians, cyclists, and intersection layouts. Across these examples, the actions remain conservative and context-aware, avoiding abrupt deviations while adapting to local traffic structure. Overall, \hydra{} demonstrates end-to-end planning robustness by maintaining smooth lane-following behavior and reasonable interaction awareness in complex urban environments.

\subsection{Additional Qualitative Results (Robotic Manipulation)}
\input{figures/viz_supp_robotics_}

Fig.~\ref{fig:viz_supp_robotics} visualizes representative frames for the ToolHang task. The sequence compares the pruned action vocabulary, shown as the top-8 candidate action chunks at each step, with the final selected action chunk executed by the policy. Across time, the candidate chunks capture diverse feasible motions for assembling the frame and lifting the tool toward the hook.

\subsection{Efficiency}
\begin{table}[h]
\centering

\caption{\textbf{Comparisons on Efficiency.} We compare the model complexity and runtime efficiency of different policies. The FPS metric is measured on one single NVIDIA A100 GPU with the PyTorch FP32 backend. * The Diffusion Policy utilizes extra networks to encode the Bird's-Eye-View condition tokens following LTF~\cite{chitta2022transfuser}, while the DDPM Scheduler~\cite{ho2020denoising} is used for iterative denoising.}
\resizebox{\textwidth}{!}{
\begin{tabular}{l|c|c|c|c|c|c}
\toprule
Method & Backbone & Params. & Vocab. Size & Iteration Steps & FPS & EPDMS \\ \midrule
Diffusion Policy*~\cite{chi2025diffusion} & V2-99 & 115.0M & - & 100 & 4.1 & 79.1 \\
Hydra-MDP-$\mathcal{V}_{16384}$~\cite{li2024hydra} & V2-99 & 83.0M & 16384 & 1 & 13.5 & 86.2 \\
GTRS-Dense~\cite{li2025generalized} & V2-99 & 83.0M & 8192 & 1 & \textbf{18.5} & 84.0 \\
ZTRS~\cite{li2025ztrs} & V2-99 & 83.4M & 8192 & 1 & 18.4 & 85.3 \\
\hydra{} & V2-99 & 82.8M & 16384 & 3 & 12.1 & \textbf{87.8} \\
\bottomrule
\end{tabular}
}
\label{tab:efficiency}
\end{table}

Tab.~\ref{tab:efficiency} compares the model complexity and runtime efficiency of different policy architectures. Compared with Diffusion Policy, \hydra{} avoids expensive iterative denoising over 100 DDPM steps, resulting in substantially higher inference speed while also improving EPDMS. For the implementation of the Diffusion Policy, we represent each trajectory as normalized displacements in the longitudinal and lateral directions. Among discrete policies, \hydra{} maintains a comparable model size with the V2-99 backbone~\cite{lee2019energy}, while using a larger action vocabulary and three lightweight selection steps. 
Although this slightly reduces FPS compared with prior one-step methods, \hydra{} still runs at 12.1 FPS on a single NVIDIA A100 GPU with the PyTorch FP32 backend, while achieving the best EPDMS score of 87.8. This indicates that the proposed multi-step action selection provides a favorable trade-off between planning accuracy and runtime efficiency.

\subsection{Broader Impact}

This work may contribute to more interpretable and physically plausible policies for autonomous driving and robotic manipulation. However, deploying such systems in the real world directly can introduce safety risks if the policy fails in unseen conditions or inherits biases from demonstration data. Practical use should therefore require rigorous closed-loop testing, safety monitoring, and human oversight.

%% file: figures/viz_supp_av_.tex
\begin{figure*}[h]
    \centering
    \includegraphics[width=\linewidth]{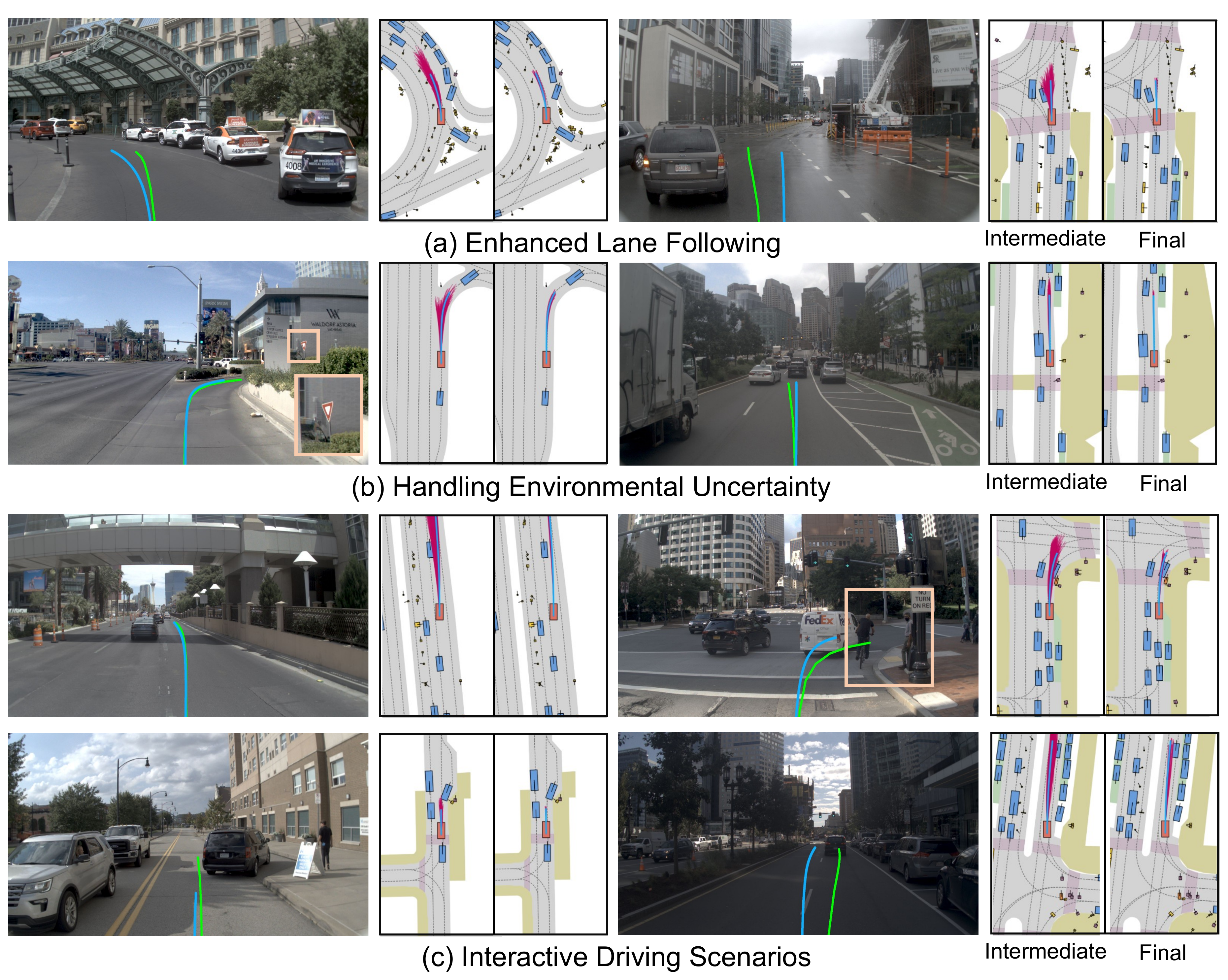}

    \caption{\textbf{Additional Visualizations on Navtest.} The figure shows the \textcolor{cyan}{selected trajectory}, the \textcolor{green}{human trajectory}, and the \textcolor{magenta}{remaining vocabulary} at the intermediate and final stages.}

    \label{fig:viz_supp_av}
\end{figure*}

%% file: figures/viz_supp_robotics_.tex
\begin{figure*}[!h]
    \centering
    \includegraphics[width=\linewidth]{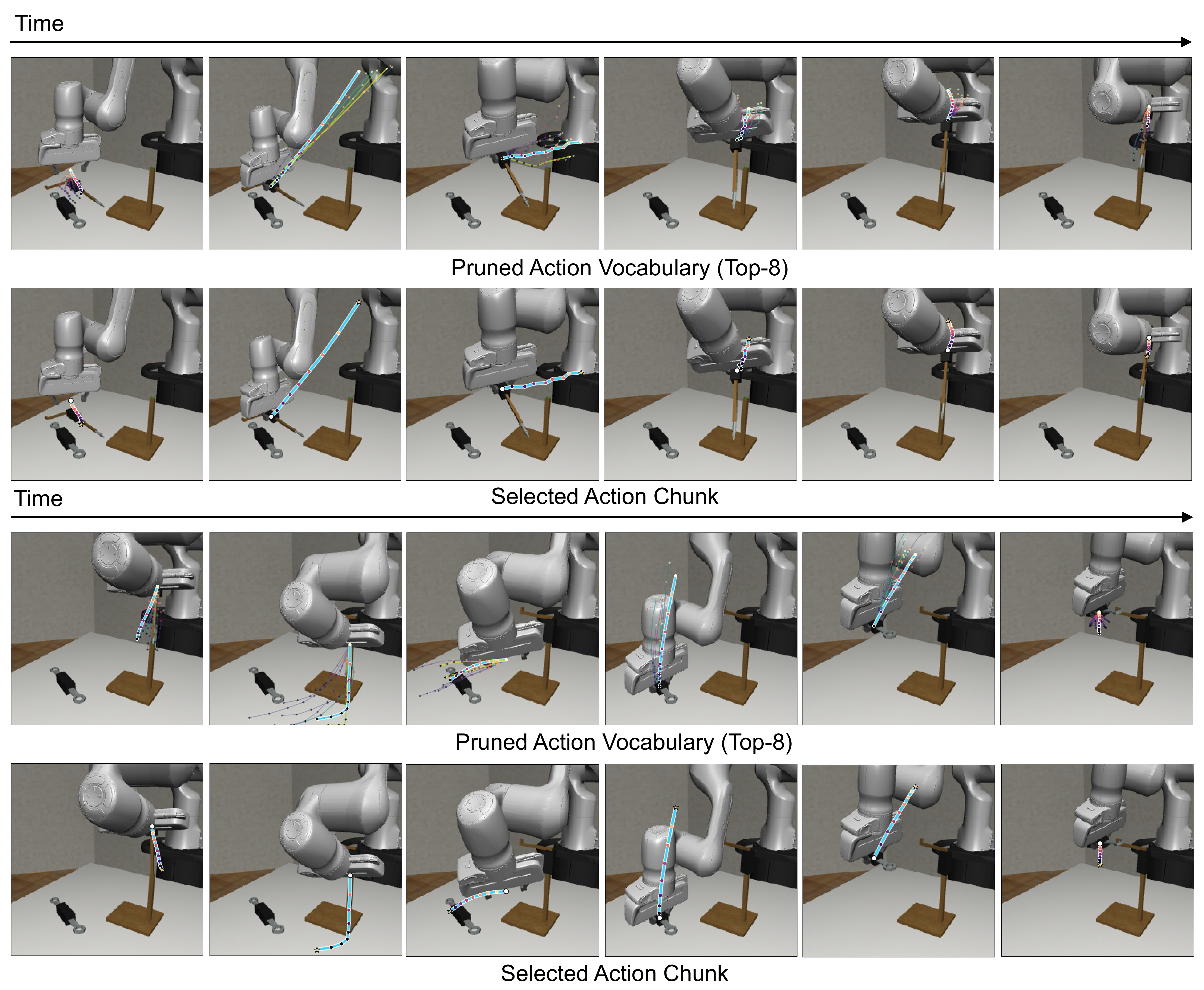}

    \caption{\textbf{Additional Visualizations on Robotic Manipulation.} The figure shows a full episode of the ToolHang task. We show the key frames with top-8 action chunks from the last scoring stage and the selected action chunk.}

    \label{fig:viz_supp_robotics}
\end{figure*}